\documentclass[10pt,twocolumn,letterpaper]{article}

\usepackage{cvpr}              

\usepackage{amssymb} 

\usepackage{pifont} 
\usepackage{graphicx}

\newcommand{\cmark}{\raisebox{-0.25em}{\includegraphics[height=1em]{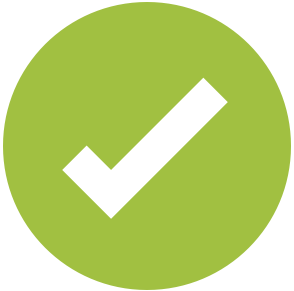}}}
\newcommand{\xmark}{\raisebox{-0.25em}{\includegraphics[height=1em]{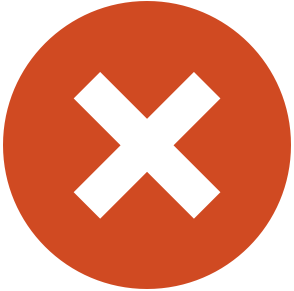}}}
\newcommand{\wmark}{\raisebox{-0.25em}{\includegraphics[height=1em]{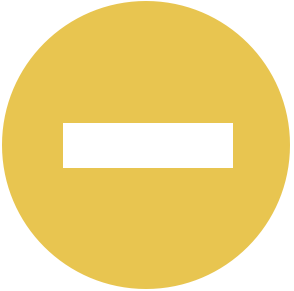}}}
\usepackage{algorithm}
\usepackage{algpseudocode}

\definecolor{cvprblue}{rgb}{0.21,0.49,0.74}
\usepackage[pagebackref,breaklinks,colorlinks,allcolors=cvprblue]{hyperref}

\usepackage{float}

\def\paperID{1} 
\def\confName{CVPR}
\def\confYear{2025}

\title{MR.NAVI: Mixed-Reality Navigation Assistant for the Visually Impaired}

\author{Nicolas Pfitzer*\\
{\tt\small npfitzer@ethz.ch}
\vspace{-5pt}
\and
Yifan Zhou*\\
{\tt\small yifzhou@ethz.ch}
\vspace{-5pt}
\and 
Marco Poggensee*\\
{\tt\small mpoggensee@ethz.ch}
\vspace{-5pt}
\and
Defne Kurtulus*\\
{\tt\small dkurtulus@ethz.ch}
\vspace{-5pt}
\and 
Bessie Dominguez-Dager\\
\vspace{-5pt}
\and
Mihai Dusmanu\\
\vspace{-5pt}
\and
Marc Pollefeys\\
\vspace{-5pt}
\and
Zuria Bauer\\
\vspace{-5pt}
}

\begin{document}
\maketitle
\begin{abstract}
Over 43 million people worldwide live with severe visual impairment \cite{blindness}, facing significant challenges in navigating unfamiliar environments. We present \textit{MR.NAVI}, a mixed reality system that enhances spatial awareness for visually impaired users through real-time scene understanding and intuitive audio feedback. Our system combines computer vision algorithms for object detection and depth estimation with natural language processing to provide contextual scene descriptions, proactive collision avoidance, and navigation instructions. The distributed architecture processes sensor data through MobileNet for object detection and employs RANSAC-based floor detection with DBSCAN clustering for obstacle avoidance. Integration with public transit APIs enables navigation with public transportation directions. Through our experiments with user studies, we evaluated both scene description and navigation functionalities in unfamiliar environments, showing promising usability and effectiveness. A demo video is available \href{https://drive.google.com/file/d/1fSBRT-kpRPz_Llu3_Z56GxEAvn-eJhMK/view?usp=sharing}{here}. Code is available \href{https://github.com/marcopepunkt/MixedReality_Baymax}{here}. 
\end{abstract}    
\section{Introduction}
\label{sec:intro}
Navigating unfamiliar environments presents significant challenges for approximately 43 million visually impaired people worldwide \cite{blindness}. Advancements in wearable devices and embedded systems, coupled with the rise of artificial intelligence (AI), are driving major innovations in assistive technologies. For individuals with visual impairments, modern solutions increasingly integrate advanced sensors and computational capabilities to enhance autonomy, overcoming the limitations of traditional aids such as white canes or guide dogs. Following the approach proposed by Lin et al. \cite{lin2017} and Brunet et al. \cite{brunet2018}, we classify these assistive technologies into three categories: \begin{itemize}[noitemsep]
    \item \textbf{Electronic Orientation Aids (EOAs)} assist users in understanding their surroundings and building a mental map.
    \item \textbf{Position Locator Devices (PLDs)} provide guidance to a final destination using GPS but lack obstacle avoidance, similar to standard navigation applications.
    \item \textbf{Electronic Travel Aids (ETAs)} focus on real-time obstacle detection and avoidance, enhancing safe mobility for users.
\end{itemize}

\begin{figure}[t!]
    \centering
    \includegraphics[width=1\linewidth]{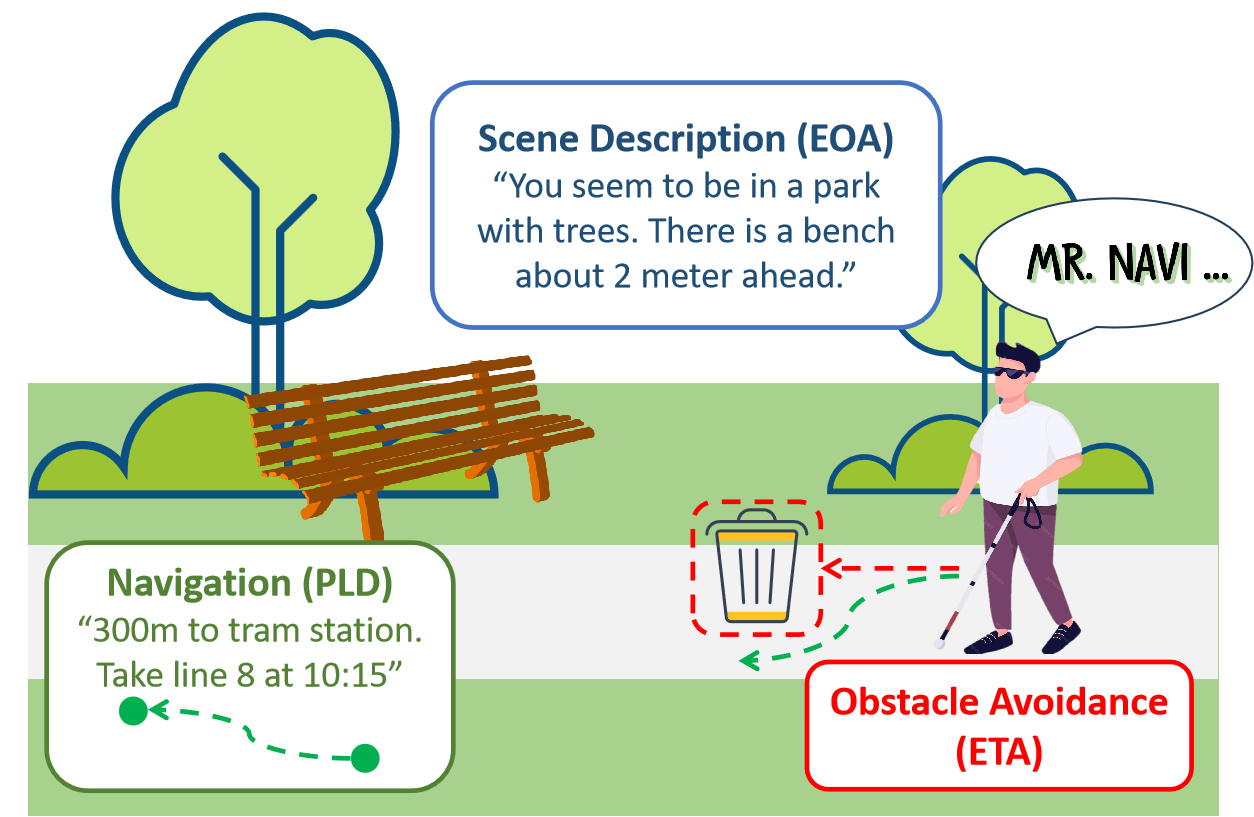}
    \vspace{-15pt}
    \caption{Our MR.NAVI system combines three assistive technologies \cite{lin2017} based on a mixed reality platform to enhance mobility for visually impaired users, showcased in a park scenario: (1) as an \textbf{Electronic Orientation Aid (EOA)}, providing contextual \textbf{scene description} information about the environment through conversations with users describing nearby objects and their distances; (2) as an \textbf{Electronic Travel Aid (ETA)}, identifying hazards and suggesting safe paths with visual indicators and spatial audio for \textbf{obstacle avoidance}; (3) and as a \textbf{Position Locator Device (PLD}), offering location-based \textbf{navigation} with public transit information.}
    \label{fig:teaser}
    \vspace{-15pt}
\end{figure}

Many assistive technologies address only a subset of these key functionalities, leaving critical gaps in usability. In addition, many solutions fall short due to impractical form factors, reliance on bespoke hardware that increases cost, and limitations in real-time scene understanding.

We present MR.NAVI, a mixed-reality (MR) navigation assistant that leverages large language models (LLMs) and computer vision to provide comprehensive, real-time spatial awareness for visually impaired users. Our system integrates real-time object detection, depth estimation, and spatial audio and uniquely addresses all three essential categories of assistive technology needs, as illustrated in Figure~\ref{fig:teaser}.
Our main contributions include:
\begin{itemize}[noitemsep]
    \item Implementation of a scene understanding pipeline (EOA) that leverages natural language processing and computer vision, optimized for real-time operation on HoloLens2.
    \item Design of an MR-based multimodal navigation system combining spatial audio, visual enhancement, and obstacle detection (ETA) with public transit guidance (PLD).
    \item Evaluation of the system's usability and effectiveness in real-world scenarios, demonstrating how the integration of all three assistive technologies enhances user confidence and navigation ability.
\end{itemize}




\section{Related Works}




For full navigation autonomy, a combination of EOAs, PLDs, and ETAs is essential. Providing only PLDs and ETAs is insufficient, as users require more than just obstacle avoidance and directional cues. In a comprehensive study of visually impaired pedestrians' needs, Brunet et al. found that visually impaired users strongly desire to form ``operative mental representations" of their environment before and during navigation—an ability enabled by EOAs. Their study showed that users who could build comprehensive mental models of the space reported higher confidence and reduced anxiety during navigation. While navigation systems often focus solely on obstacle avoidance and basic directions, the study found that the users need ``detailed information concerning various aspects of the map" including rich descriptions of street layouts, crossing points, and environmental context and highlight the importance of EOA technologies to ``form a clearer mental representation of the itinerary and to better anticipate difficulties"~\cite{brunet2018}.

Several promising approaches have emerged, each addressing specific aspects of navigation. These approaches are compared in Table \ref{tab:comparison}. The feasibility of smart glasses for obstacle avoidance was demonstrated by Bai et al.~\cite{bai2017smart}, showing that computer vision techniques can be used to help the user avoid obstacles and find a suitable path, with a spatial audio system. Similarly, Schwarze et al. \cite{schwarze2016} proposed an obstacle avoidance system that generates bounding boxes around dynamic objects and estimates their likely trajectories, providing users with information about obstacles through distinct audio cues. More recently, the research startup \textit{.Lumen}~\cite{dotlumen2024} developed a dedicated assistive technology, which combines computer vision with haptic and audio feedback to provide obstacle detection and navigation assistance. However, while these solutions excel at providing directional guidance, they still lack the essential EOA and PLD features needed to help users form a rich mental representation of their environment and a long distance navigation. 

Other works have started to address the need for stronger EOA capabilities by integrating them into ETA technologies. Madake et al.~\cite{madake2023} proposed a wearable assistive system that captures images and generates the corresponding verbal descriptions of the environment. The ETA is provided through a vest that gives haptic feedback to the user when a collision is imminent. However, PLD is not integrated into the system.
Beyond obstacle avoidance (ETA) and environment awareness (EOA), PLDs focus on longer distance navigation, but remain separated from ETA and EOA systems. Two commonly used applications are the mobile Applications Lazarillo~\cite{lazarillo} and BlindSquare~\cite{blindsqare}. They extend traditional GPS-based navigation services by making them accessible to visually impaired users through verbal guidance. These PLDs do not help the user to avoid obstacles or build a mental representation of the environment.

One approach to combine all three technologies is ISANA~
\cite{isana2016}. It provides indoor guidance through a combination of location context awareness and obstacle detection. While ISANA offers all functionalities required for full navigation, it relies on a detailed map of the indoor environment, limiting its adaptability across different scenarios and dynamic environments.

\begin{table}[t]
    \centering
    \caption{\textbf{Comparison of related works.} Comparison of current solutions in academia and industry regarding the EOA (Electronic Orientation Aids), PLD (Position Locator Devices) and ETA (Electronic Travel Aids) functionality. (\cmark~--~fulfilled, \xmark~--~not fulfilled, \wmark~--~partially fulfilled)} 
    \vspace{-5pt}

    \begin{tabular}{cccc}
        \toprule
        \textbf{Approach}  & \textbf{EOA} & \textbf{PLD} & \textbf{ETA}  \\
        \midrule
        Bai et al. \cite{bai2017smart} & \xmark & \xmark & \cmark  \\
        Schwarze et al. \cite{schwarze2016} & \xmark & \xmark & \cmark \\
        .Lumen \cite{dotlumen2024} & \xmark & \xmark & \cmark \\
        Madake et al. \cite{madake2023} & \cmark & \xmark & \cmark \\
        Navigation apps \cite{lazarillo}\cite{blindsqare} & \xmark & \cmark & \xmark \\
        ISANA \cite{isana2016} & \cmark & \wmark & \cmark \\
        \midrule
        \textbf{MR.NAVI (ours)} & \cmark & \cmark & \cmark \\
        \bottomrule
    \end{tabular}
    \label{tab:comparison}
    \vspace{-10pt}
\end{table}




Building upon these foundations, our work integrates key strengths from existing solutions while addressing their limitations in a commercially available and affordable package. The limitations are addressed through the integration of multiple feedback modalities, real-time processing, and natural language interaction. By combining advanced computer vision with conversational AI capabilities, we aim to create a system that provides EOA, PLD, and ETA functionalities.

\begin{figure*}[ht!]
  \centering
  \includegraphics[width=1\textwidth]{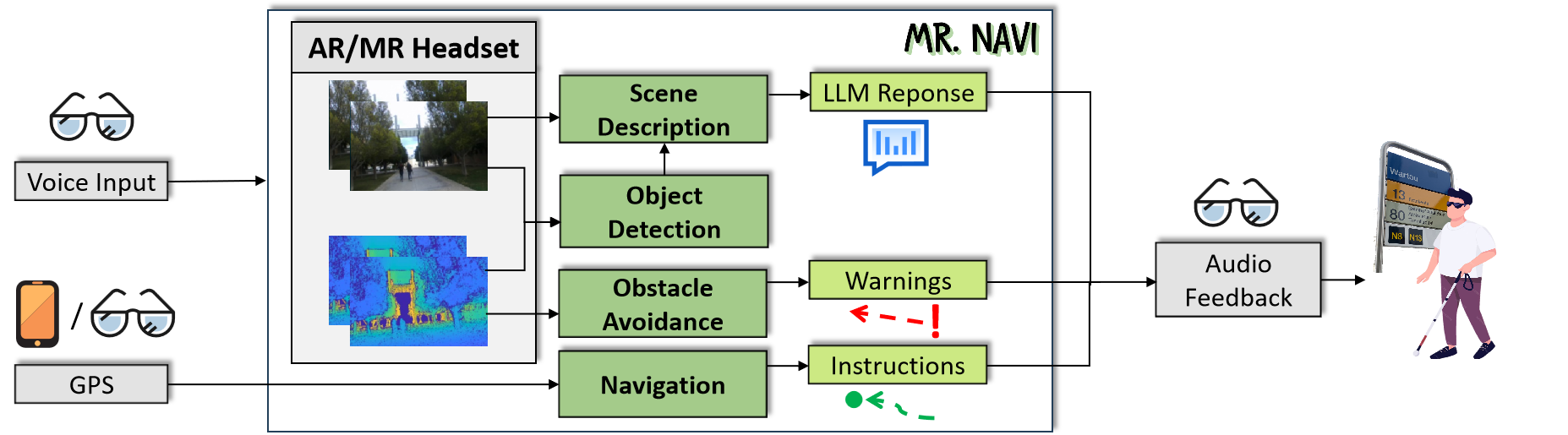}
  \vspace{-10pt} 
     \caption{The system architecture illustrating the complete data flow and component interactions. The system processes input from user voice commands, AR/MR headset sensors (camera images and depth data), and GPS to deliver audio feedback to visually impaired users through four key functional components: Scene Description, Object Detection (identifying relevant objects in the user's surroundings), Obstacle Avoidance (providing real-time warnings about potential hazards), and Navigation (offering directional guidance using GPS and map data). This integrated approach enables users to understand their environment, avoid collisions, and navigate safely to their destinations. The pipeline shown in the figure illustrates the application with a generic MR headset, while our specific prototype implementation utilized the HoloLens2 with research mode sensor streaming \cite{hololens2} to access the required sensor data.}
  \label{fig:pipeline}
  \vspace{-10pt} 
\end{figure*}

\section{App Framework}

The \textit{MR.NAVI} system comprises four main functional components that work together to assist visually impaired users in navigating urban environments while providing them with an understanding of their unfamiliar surroundings: Scene Description, Object Detection, Obstacle Avoidance, and Navigation. 

To support this functionality, the full data flow and component interactions of our system are illustrated in Figure \ref{fig:pipeline}. This pipeline shows how voice input, sensor data of an AR/MR headset, and four functional modules are integrated into our application.

\subsection{Scene Description}
The user can pose any question about the scene in front of them, and the app will provide concise verbal descriptions, identifying key environmental objects and their distances to the user. Follow-up questions can be asked within the same conversation, based on a single camera frame.

\subsection{Obstacle avoidance}
\begin{figure}[t]
  \centering
  \includegraphics[width=.95\linewidth]{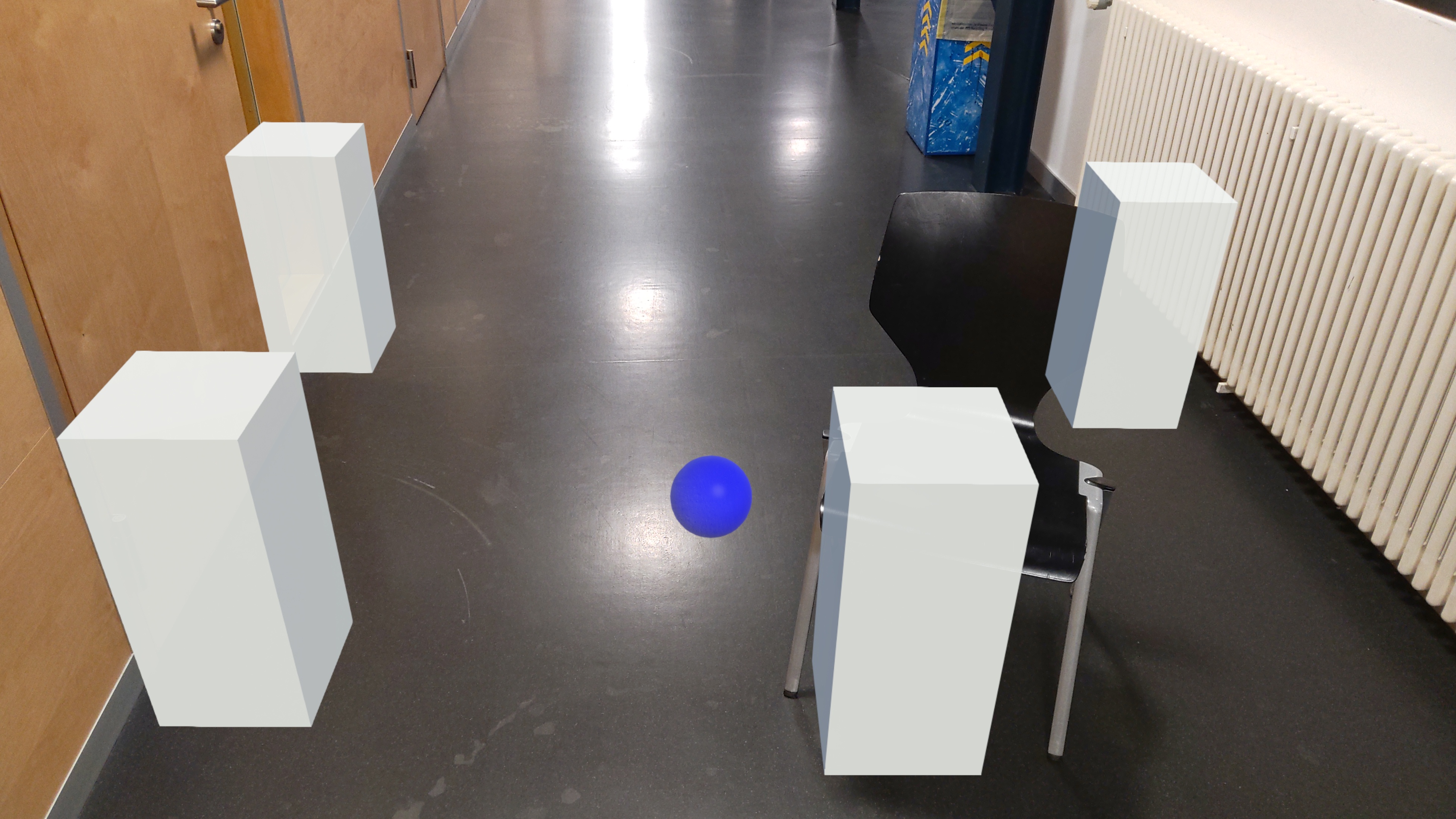}
  \vspace{-5pt}
    \caption{\textbf{Obstacle Avoidance:} a visual representation of the obstacle detection and avoidance system. White cubes indicate detected obstacles (including walls). The blue sphere visualizes the suggested safe path direction calculated by the obstacle avoidance algorithm (see Section \ref{obstacleAvoidance}). This visual guidance is complemented by spatial audio cues coming from the direction of the blue sphere to assist users with limited or no vision.}
   \label{fig:directions.jpg}
   \vspace{-13pt} 
\end{figure}

Our system provides the user with a multimodal feedback system that combines visual enhancement and spatial audio feedback. The real-time processing pipeline continuously analyzes depth sensor data to maintain accurate obstacle positions and dynamically updates the navigation path. For users with partial vision, we render detected obstacles as white cubes, while a floating blue sphere indicates the recommended safe path direction, as illustrated in Figure \ref{fig:directions.jpg}. This visual guidance is complemented by a spatial audio system that provides directional cues through binaurally rendered sounds, in the direction of the obstacle-free path.

\subsection{Google Maps based navigation}

Users can request guidance to a specific address, and the system provides through auditory feedback transit instructions based on their current location and real-time public transport data provided by the Google Maps API \cite{google_maps_api_python}, as illustrated in Figure \ref{fig:maps_directions.jpg}. The directions provide the user information about tram or bus lines they need to take, along with departure stop and times. 

\begin{figure}[t!]
  \centering
  \includegraphics[width=0.95\linewidth]{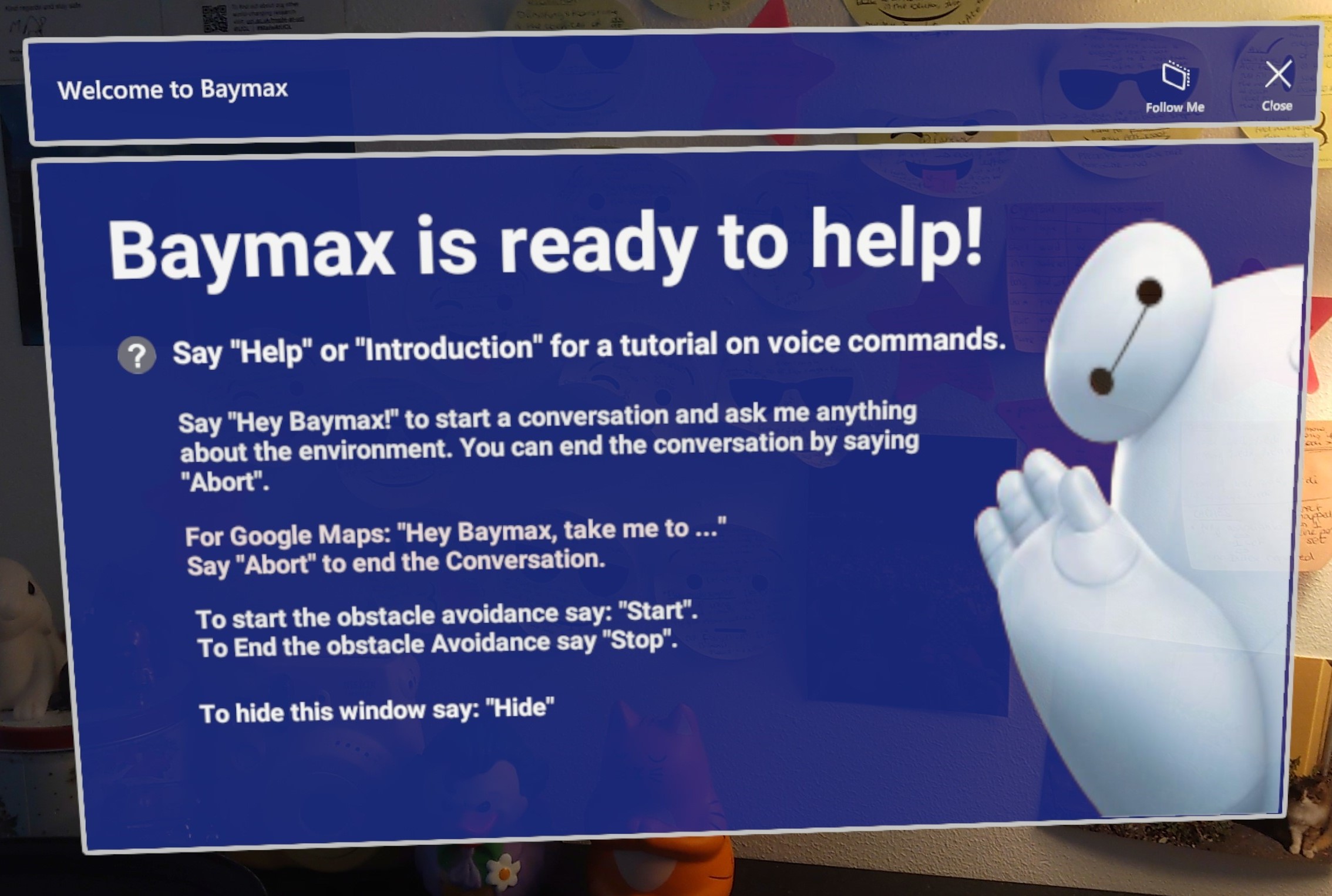}
   \caption{\textbf{Welcome Page:} the user interface of the navigation assistant for visually impaired users, showing the welcome screen with voice command instructions. The screen displays available voice commands for scene description, navigation assistance, and obstacle avoidance. Note that the interface shown uses a robot character name and appearance only for demostration purpose.}
   \label{fig:homescreen.jpg}
   \vspace{-5pt} 
\end{figure} 

\subsection{User Interface}
\begin{figure}[t]
  \centering
  \includegraphics[width=1\linewidth, trim = 4cm 2cm 6.5cm 4cm, clip]{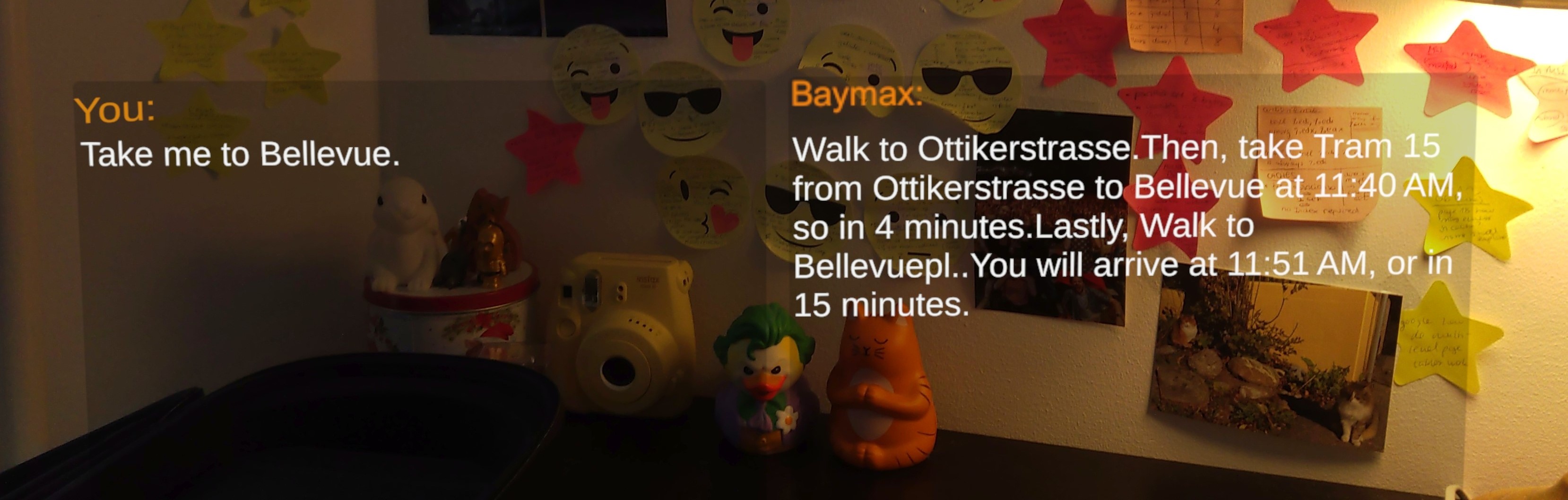}
  \vspace{-15pt}
      \caption{\textbf{Navigation:} the navigation interface displaying transit guidance retrieved from the Maps API in response to a user's voice command. The system processes voice input (e.g., ``take me to [destination]") and visualizes it as on-screen captions, showing the user's request alongside the response. Detailed information about available public transportation options is provided, including bus/tram lines, departure stops, and timing information. This dual-mode presentation - visual captions and verbalized audio feedback - ensures accessibility for users with varying levels of visual impairment.}

   \label{fig:maps_directions.jpg}
   \vspace{-10pt} 
\end{figure}

The primary interface enables hands-free control through voice commands. The welcome page shown on Figure \ref{fig:homescreen.jpg} illustrates the components of the user interface, along with an audio tutorial. In our prototype implementation and demonstration video, we used an assistant name and visual appearance inspired by the robot assistant character \textit{Baymax} in \textit{Big Hero 6} (Walt Disney Animation Studios, 2014), purely for demonstration purposes. For the actual system implementation, we use "\textit{Hey Mr.Navi}" as the default wake word, though this remains customizable to accommodate user preferences. After the wake word, specific commands can be issued.  Saying \textit{"help"} triggers verbal instructions for using the system. For users with partial vision, the system provides visual feedback through an interface displaying detected objects and navigation cues (Figure \ref{fig:directions.jpg}), as well as \textit{MR.NAVI}'s reply to the user's queries (Figure \ref{fig:maps_directions.jpg}). 

\section{Implementation}
\label{implementation}

The system architecture comprises two primary modules. While most user interactions are processed directly on the HoloLens2 device, computationally intensive tasks are offloaded to a backend server, which maintains continuous communication with the headset.

Communication between the HoloLens2 and the PC server, including the transmission of user input and the results of algorithmic and API-based processes, is handled via HTTP requests and JSON packets. Sensor data from the HoloLens2, including RGB and depth frames, is streamed to the PC using the HoloLens2 Sensor Streaming framework \textit{hl2ss} \cite{hl2ss}. The PC server also receives GPS coordinates from the user's phone, used for the Google Maps feature. Most user interactions are handled through voice commands and audio feedback and all speech related communications are powered by Azure Speech Services \cite{azure_speech}. The app’s graphical user interface was developed using the Unity game engine \cite{unity}.

For seamless communication while running the app, the HoloLens2, PC, and phone are connected to the same Wi-Fi network. To meet user requirements for low latency, we recommend using a reliable and high-speed Wi-Fi connection, possibly through a router. As future implementations shift more of the computational load onto the device itself, this dependency on network performance is expected to decrease, making latency and connectivity less of a concern.

The implementation of each feature in our app is described in the following subsections.

\subsection{Scene Description}
\label{sceneDescription}

When the user activates the scene description feature, their voice query, along with the current RGB and depth frames from the HoloLens2, is transmitted to the HTTP server on the PC.

For each new frame, object detection is performed first to identify nearby objects, along with their corresponding depths. This information is used to potentially alert the user to objects in close proximity and their distances, which can be later integrated into the prompt with the image frame to the chatbot. We utilize MobileNet \cite{mobilenet} for object detection due to its compatibility with the libraries in our implementation and its ability to achieve sufficiently fast runtimes on a CPU environment.

The depth estimation is performed before querying the Gemini API, as large language models lack the capability to accurately estimate depth from a single RGB image. Although we considered monocular depth estimation models, such as UniDepth \cite{unidepth}, we opted to use HoloLens2 long-throw depth sensors due to their superior reliability and seamless integration into our pipeline.

The detected objects and their estimated depths, along with the user’s query, are then passed to the Gemini API \cite{gemini_api}. We chose to consult a chatbot for the scene description because it identifies a broader range of objects compared to our object detection model and provides responses in a user-friendly language, which is essential for our assistive application.

Finally, the chatbot’s scene description is sent back to the HoloLens2 via a JSON packet and then announced to the user.


\subsection{Obstacle avoidance}
\label{obstacleAvoidance}

For obstacle avoidance, real-time depth data from the HoloLens2 Longthrow sensor is streamed to the PC server along with head-tracking information. Each incoming depth frame is converted into a 3D point cloud, which is then registered to a global reference frame (GRF) using the measurements of the user’s head pose and an initial position fixed upon the launch of the app. To maintain near real-time performance (close to the 5 fps limit of the depth sensor), the point cloud is first downsampled using a voxelized grid approach. Performing obstacle avoidance in the GRF offers two key advantages:
\begin{itemize}
    \item \textbf{Obstacle Tracking}: The system maintains a dynamic list of previously detected obstacles on the server, enabling reliable obstacle avoidance by incorporating scene elements beyond the device’s direct field of view. Additionally, by merging multiple observations of the same obstacle from different viewpoints, the system refines position estimates, reducing localization errors. This significantly enhances robustness, particularly in large and cluttered environments.
    
    \item   \textbf{Floor Detection}: Raw point clouds do not inherently differentiate floor points from walls, ceilings, or other obstacles. To isolate actual obstacles, the floor must be removed. In the GRF, this can be achieved by fitting a horizontal plane using RANSAC \cite{ransac}, which effectively filters out the floor. In the local (head-centered) reference frame, the user’s head tilt would complicate this filtering, since the system would have no fixed notion of what constitutes a horizontal plane. 
\end{itemize}

\begin{figure}[h]
  \centering
  \includegraphics[width=0.8\linewidth]{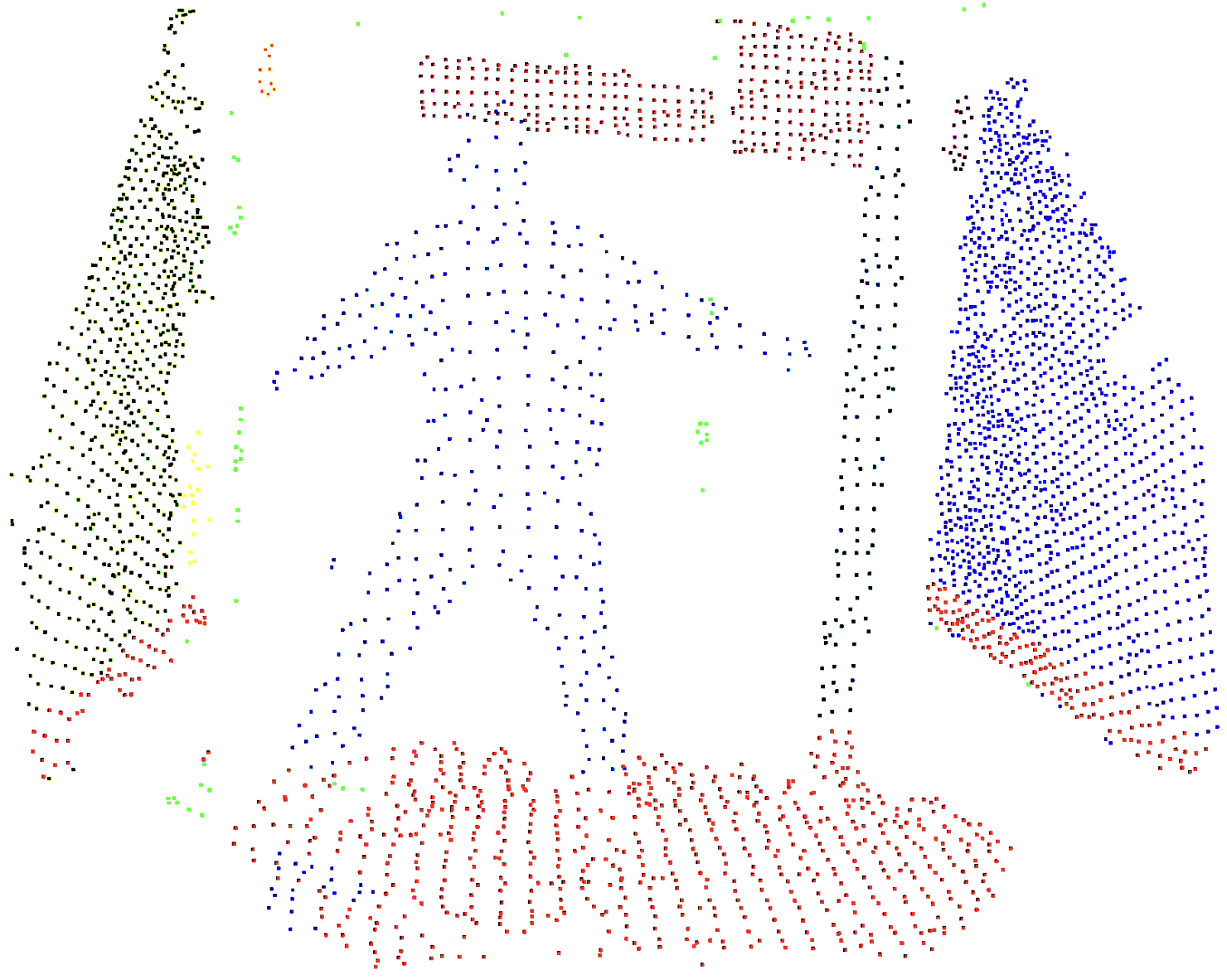}
  \vspace{-10pt}
      \caption{\textbf{Point Cloud Segmentation:} a visualization of the environmental mapping process that enables obstacle detection. We observe a person blocking the way in a narrow corridor. The floor (shown in red) is identified using RANSAC and filtered out before the clustering process. Each remaining cluster of points represents a potential obstacle and is assigned a distinct color for visualization purposes, with light green points representing outliers.}
   \label{fig:pointcloud.png}
   \vspace{-10pt} 
\end{figure}

An illustration of this process, including point cloud segmentation and obstacle detection, is shown in Figure \ref{fig:pointcloud.png}.

After the floor has been identified and removed, the remaining points in the global point cloud are segmented into clusters using the Density-Based Spatial Clustering of Applications with Noise (DBSCAN) algorithm \cite{dbscan}. This segmentation is essential for the system to differentiate between navigable spaces and obstacles. DBSCAN is particularly well-suited for this scenario because it does not require prior knowledge of the number of clusters, and it can easily separate sparse outliers from meaningful groups of points. Once the points have been grouped, each cluster is enclosed in a bounding box, creating a compact geometric representation of the obstacle. These bounding boxes are then used for collision avoidance.

\begin{figure}[h]
  \centering
  \includegraphics[width=\linewidth]{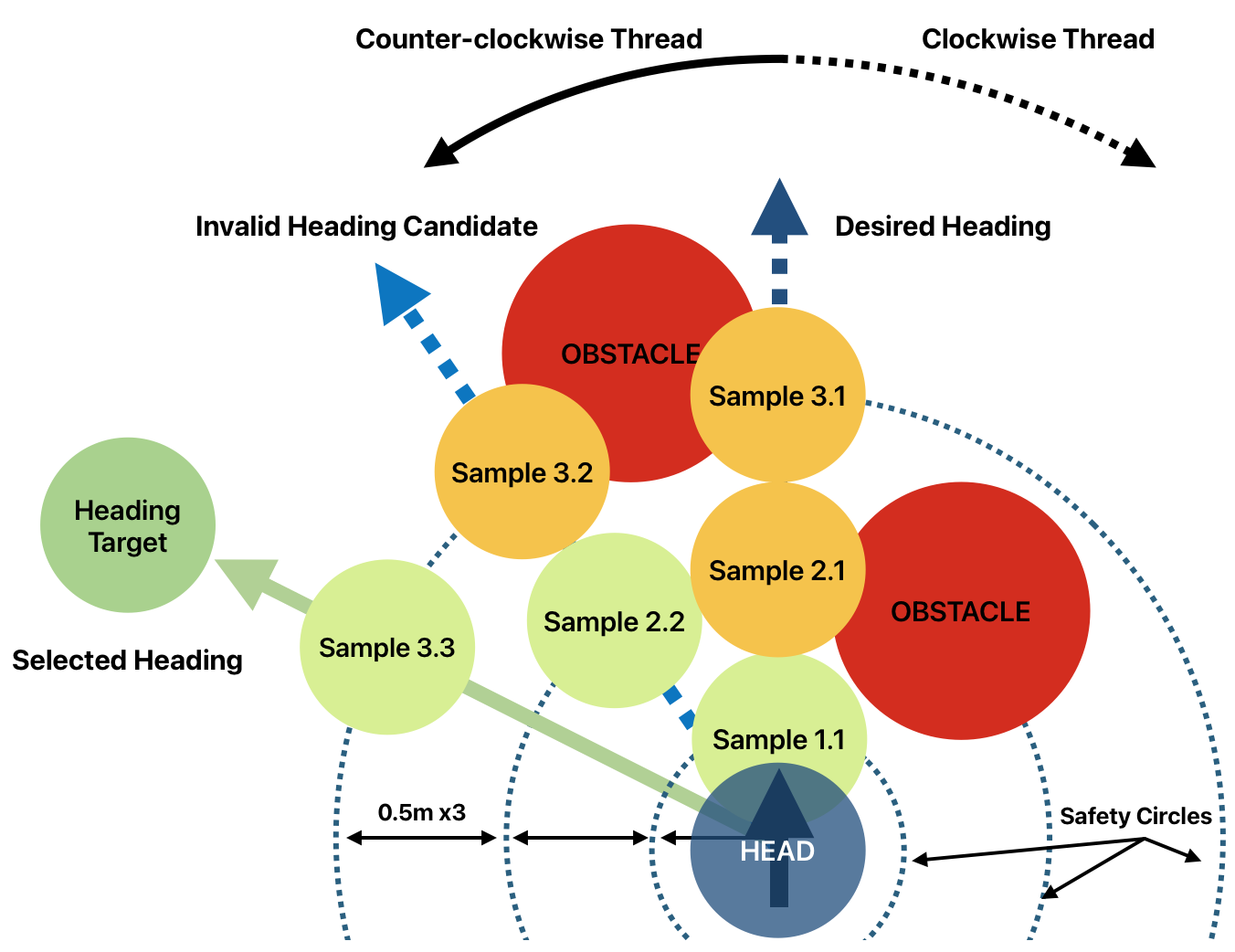}
  \vspace{-15pt} 
  \caption{\textbf{Heading diagram:} a technical illustration of the path-finding algorithm used for obstacle avoidance. The diagram illustrates how the algorithm samples potential directions around the user’s current position (labeled “HEAD”) by expanding outwards through multiple safety circles (shown as dashed blue lines). Each colored circle (e.g., “Sample 1.1,” “Sample 2.2,” “Sample 3.2”) represents a candidate heading at that distance, with orange samples indicating invalid paths (too close to obstacles) and green samples showing valid paths with sufficient clearance. Larger red circles represent detected obstacles. The algorithm ultimately returns the "Heading Target", which is the valid heading with the smallest absolute angle to the "Desired Heading". The algorithm is further accelerated by parallelizing clockwise and counter-clockwise searches.}
  \label{fig:heading.png}
  \vspace{-10pt} 
\end{figure}

A collision-free heading is computed on a 2D projection of the stored scene, which includes both currently observable obstacles and memorized ones. In essence, the algorithm systematically evaluates possible orientations in an expanding ring around the user, rapidly converging on the angle that offers the most obstacle-free path. Refer to Figure \ref{fig:heading.png} for an illustration of the path-finding process.

\subsection{Google maps based navigation}
\label{google_maps_impl}

For the Google Maps-based navigation feature, the user's desired destination, along with their GPS coordinates from their phone, are sent to the PC server. The GPS coordinates are transmitted via the GPS2IP Lite \cite{gps2ip} app.

The Google Maps Directions API \cite{google_maps_api_python} in Python returns relevant public transportation details to the destination — such as lines, times, and departure stops — along with GPS coordinates for the stops. This data is transmitted to the HoloLens2 app and the instructions are played through audio to the user.

Upon request, the app can also provide walking directions to the next public transportation stop by continuously comparing the user's current GPS coordinates with those of the next waypoint, triggering instructions at appropriate intervals. These walking instructions specify the walking direction along with distance and time (in meters and minutes, respectively). However, the instructions currently provided by the Google Maps API often lack sufficient clarity for visually impaired users—for example, offering vague directions such as "Walk south-east for 10 meters," which can be confusing even for sighted individuals. Addressing these limitations is essential and is discussed further in Section \ref{section_google_maps_future}.

\section{User Study}

User studies were conducted to evaluate \textit{MR.NAVI}. Fourteen participants, none of whom had significant visual impairments, first evaluated the visual interface under normal conditions. They were then asked to wear a blindfolded during testing to simulate how a visually impaired user might experience the app.

The aim of the user studies was to gather both qualitative and quantitative insights into the app’s performance, addressing the following research questions:

\begin{itemize}[noitemsep]
    \item Does the app provide useful information about the surroundings for a visually impaired user?
    \item Does the app deliver accurate auditory and visual cues for the obstacle avoidance navigation mode?
    \item Is the app easy and intuitive to use, considering the user interface and spatial audio features?
\end{itemize}

Quantitative metrics were employed to assess the response time of the scene description, as well as the accuracy of the obstacles' positions and the safe navigation path generated by the obstacle avoidance mode. Qualitative feedback was collected to evaluate user experience and identify areas for improvement to determine how the app’s features can be enhanced.

\subsection{Experiment Settings}

All user studies were performed indoors within university buildings. During the evaluation, participants were asked to perform different tasks for each feature of the app and then complete a survey to help us answer our research questions. None of the participants had prior experience with the HoloLens2 or any mixed-reality headset. 

Additionally, the app was tested outdoors by the developers; further details on these results are provided in Section \ref{outdoors}.

\subsubsection{Tasks}
\label{tasks}

Each participant was first provided with a tutorial on how to use the available voice commands, which are also outlined on the app's welcome page.

For the scene description feature, participants were instructed to make the following queries to \textit{MR.NAVI}:

\begin{itemize}[noitemsep]
    \item Ask what is in front of them.
    \item Read text from signs, such as auditorium names with floors, store opening hours, restroom signs, cafeteria names and vending machines.
    \item Request navigation instructions, such as ``How do I get out of the building or this hallway?"
\end{itemize}

For the obstacle avoidance mode, participants first walked through a hallway with obstacles such as chairs and walls while keeping their eyes open, allowing them to see the visual cues on the user interface. Then, they were asked to repeat the task whilst wearing a blindfold and follow the spatial audio cues. During this phase, additional moving obstacles were introduced in front of the participants to simulate real-world conditions and evaluate their ability to navigate based solely on auditory feedback.

For the Google Maps-based navigation feature, participants could request transit directions to a destination address, along with walking directions to the first relevant waypoint on the route, such as a public transport stop or a road crossing.

\subsubsection{User survey}

We asked 16 questions in a 
Google Form
 to the users about their experience. While some questions were omitted for brevity, the key questions included are as follows:


\begin{enumerate}[noitemsep] 
    \item Did the scene descriptions provide you with useful information about your surroundings? (rate from 1 to 5)
    \item Is there any other kind of information you would have liked to get from the scene description?
    \item Did the obstacle avoidance help you feel more confident navigating the environment? (rate from 1 to 5)
    \item Were the obstacle warnings timely and accurate?  (rate from 1 to 5)
    \item Was the spatial audio helpful in avoiding obstacles? (rate from 1 to 5)
    \item Do you think other kinds of sound effects or visual enhancements would be more helpful? 
    \item Is there any other kind of information you would have liked to get from Google Maps?
    \item How easy was it to interact with the app using the available commands? (rate from 1 to 5)
    \item Are there any other features you would have liked, to feel more comfortable while navigating blindly?
    \item If you were visually impaired, would you use the app? (rate from 1 to 5)
\end{enumerate}

We then gathered the ratings and feedback for evaluation and future improvements.

\subsection{Results, Limitations and Future Improvements}
\label{section_results}

For each feature, we evaluated both its accuracy and usefulness, gaining valuable insights into its limitations and identifying potential solutions.

\subsubsection{Scene Description}

The feature received positive feedback from all participants who tested it and was regarded as a helpful tool. Additionally, 85\% of the participants considered the length of the chatbot’s scene descriptions to be appropriate. In terms of content, users expressed high satisfaction, giving it an average rating of 4.25 out of 5. The app successfully performed all tasks outlined in Section \ref{tasks}. 

Across 20 tests conducted by the \textit{MR.NAVI} developers, while connected to a router, the scene description demonstrated a mean response time of 5.5 seconds, with a standard deviation of 5.1 seconds. However, some participants of the study reported longer delays. This difference in response time can be attributed to longer and more complex scene descriptions, as well as the presence of numerous Wi-Fi networks in close proximity, causing interference and reducing the reliability of the connection in the test location. A stable connection and continuous token streaming could mitigate these effects. Additionally, some users noted the limited ability of the scene description framework to recognize and emphasize moving objects, identifying this as a potential area for improvement. 

\subsubsection{Obstacle avoidance}

\begin{figure*}[h]
  \vspace{-14pt}
  \centering
  \includegraphics[width=0.95\linewidth]{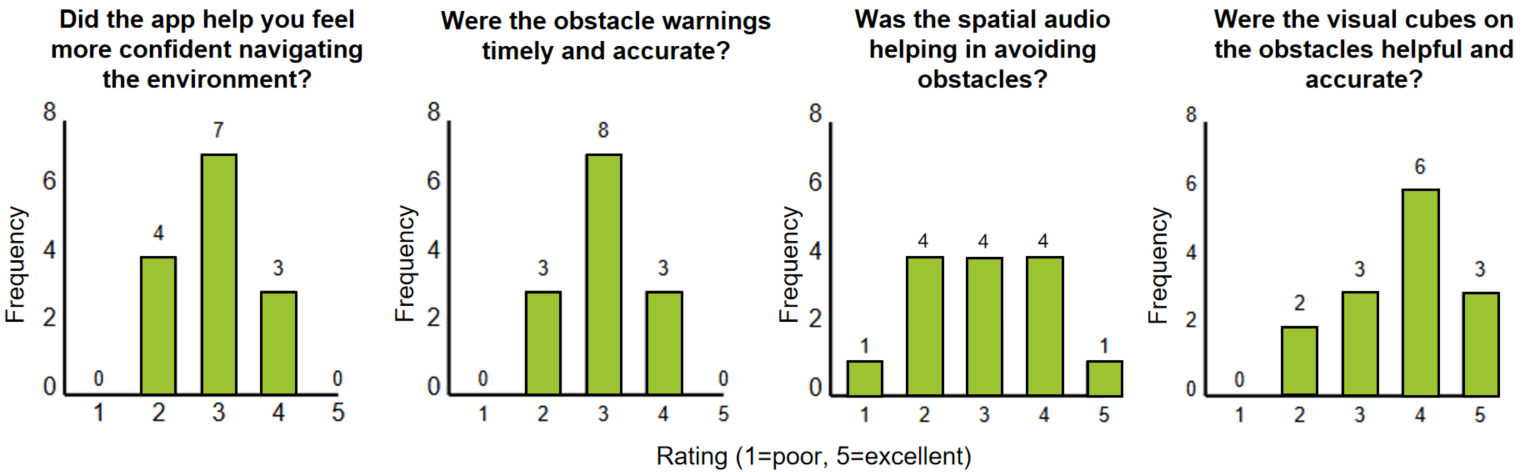}
  \vspace{-5pt}
    \caption{\textbf{User Studies Feedback:} Quantitative results from user evaluation of the obstacle avoidance feature. The chart displays participant ratings across four key assessment criteria: confidence in navigation, timeliness of obstacle warnings, effectiveness of spatial audio cues, and accuracy of visual obstacle representation. Each criterion was evaluated on a 5-point scale (1=poor, 5=excellent), with frequency distributions showing varied user experiences. Most notably, obstacle warnings received the highest overall satisfaction (8 participants rating it 3/5), while spatial audio feedback showed the most uniform distribution of ratings, indicating inconsistent user experiences with this modality.}

   \label{fig:feedback_obstacle_avoidance}
   \vspace{-10pt} 
\end{figure*}

The obstacle avoidance mode received mixed ratings and feedback from the 14 participants who tested it. While some aspects were positively reviewed,  particularly the timeliness of obstacle warnings with 8 participants rating it 3/5, other components revealed significant opportunities for improvement. Users specifically highlighted limitations in the auditory feedback system and the behavior of the guidance ball that indicated safe paths. The visual representation of obstacles received generally favorable ratings, with 6 participants rating it 4/5, though spatial audio cues showed more distributed responses across all rating levels. These quantitative assessments, summarized in Figure \ref{fig:feedback_obstacle_avoidance}, provide critical insights for future refinements of both the system's accuracy and usability.

\textbf{Visual cues.} The cubes representing obstacles were generally accurately positioned, achieving an average user rating of 3.71 out of 5. Most obstacles are successfully detected, with the exception of glass walls and stairs. While the guiding ball often avoided obstacles as intended, it could sometimes change place abruptly or appear outside the user’s field of view. The ball may direct users towards obstacles located more than 2 meters ahead of them, as it updates to point toward the free path 1.5 meters ahead of the user. Once the user nears the obstacle, the ball adjusts direction to continue guiding them safely. Future implementations could extend the safe path calculation to cover a greater distance. Planning algorithms such as Rapidly Exploring Random Trees (RRT) \cite{RRT} or Dijkstra’s Algorithm \cite{dijkstra} could help users anticipate distant obstacles earlier in their journey. In addition, the guiding ball can be replaced with a more intuitive shape, such as an arrow, to enhance the user interface.

\textbf{Auditory cues.} The feedback on spatial audio varies across users, with an average rating of 3 out of 5 for ease of use. Many users reported difficulty in determining the obstacle-free path based solely on stereo audio cues when their eyes were closed. A consistent observation was that the HoloLens2' spatial audio cues for right and left directions were generally easy to follow, whereas interpreting forward and backward directions was more challenging, as the sound seemed to come from both sides of the HoloLens2 simultaneously. While we implemented the audio to originate from the guidance sphere's 3D spatial coordinates in our Unity application, the HoloLens2 speaker system faces hardware limitations in delivering precise spatial audio perception. 
Future enhancements to convey more precise directional information may include playing distinct sounds for each direction, providing verbal cues, or integrating the application with haptic feedback systems.


\textbf{Alternative implementation.} User feedback indicated that the continuous audio cues might overwhelm the user, particularly those with visual impairments who often have heightened auditory sensitivity. A more user friendly navigation mode could warn users of immediate obstacles through sound cues, instead of playing constant audio. The app could also verbally inform the user when there are no obstacles within a certain radius. As obstacles approach, the sound could increase in intensity or change frequency, depending on the proximity of the obstacle.

\subsubsection{Google Maps based navigation}
\label{section_google_maps_future}
As mentioned in Section \ref{google_maps_impl}, the walking instruction feature is challenging to evaluate due to the vagueness of certain directions. Nonetheless, users were able to request transit routes to specific destinations such as a university or a train station and found the information helpful for transportation planning.

A key limitation is that the waypoints generated by Google Maps are too sparse and not designed with the needs of visually impaired users in mind. A future implementation could address this by sampling a denser set of waypoints along the route, enabling more continuous and context-aware navigation in parallel with the existing obstacle avoidance system. Additionally, integrating alternatives to Google Maps such as Lazarillo \cite{lazarillo}, which is specifically developed to compute safer routes for visually impaired individuals, could significantly enhance the system’s overall accessibility and usability.

\subsubsection{User Interface}

The user interface was generally regarded as easy to use, earning an average rating of 3.93 out of 5 for usability by 14 users. There were occasional issues with speech recognition in noisy environments, where commands such as ``start" or ``stop" were not recognized. Additionally, a poor Wi-Fi connection can lead to prolonged loading times for the scene description, leaving the user unsure about how to interact with the interface. This could be addressed by implementing a timeout that returns the user to the welcome page with an audio prompt, or by ensuring a stable Wi-Fi connection.

Additionally, the task of starting the app and server on a PC -- including configuring the PC’s IP address on the HoloLens2 app -- is not feasible for blind users. The app can be launched via Microsoft’s Cortana, but external assistance may still be required to initialize the server on the PC. Future work could explore implementing on-device processing directly on the HoloLens2, thus eliminating the need for a separate back-end server.

\subsubsection{Outdoors testing}
\label{outdoors}
Testing the app outdoors revealed several operational constraints. The HoloLens2 camera demonstrated significant limitations in varying lighting conditions, particularly during low-light situations (e.g., after sunset), which severely impacted the system's object detection capabilities. The HoloLens2 depth sensor, like most commercial depth sensors, faces well-documented challenges with long-distance measurements and strong sunlight conditions \cite{depthsensor}. The camera's narrow field of view further restricted its ability to capture comprehensive environmental information. Network connectivity posed another challenge, as the system's reliance on WiFi resulted in occasional latency issues and connection drops outdoors. Finally, we observed performance degradation in cold weather, with the device experiencing increased lag and reduced responsiveness. 

Future iterations of the system would benefit from implementing robust monocular depth estimation algorithms such as UniDepth~\cite{unidepth} alongside sensor fusion techniques to combine the strengths of both methods, particularly in challenging lighting conditions. 

\section{Conclusion}

In this work, we presented \textit{MR.NAVI}, a mixed reality navigation assistant designed to enhance spatial awareness for visually impaired individuals using HoloLens2, through real-time obstacle avoidance and natural language-based scene descriptions.

The user study participants, when asked whether they would use \textit{MR.NAVI} if they had visual impairments, gave an average rating of 3.79 out of 5.0. Based on this feedback, it is reasonable to suggest that our proof-of-concept can deliver useful environmental information and contribute to a sense of safety, especially when used alongside other assistive tools such as a cane or haptic devices.

A key insight from our evaluation is the need for more intuitive spatial audio cues and extended safe-path planning to improve real-time guidance. Future work will focus on refining the navigation algorithm to cover longer distances, integrating event-triggered auditory feedback to reduce cognitive load, and improving robustness in outdoor environments through sensor fusion techniques. Moreover, further validation with visually impaired users will be essential to tweak the user experience and assess real-world usability.

By combining computer vision, spatial audio, and natural language-driven scene understanding, \textit{MR.NAVI} represents a step toward more intelligent and accessible mixed reality solutions for the visually impaired. With continued advancements, we believe our system can contribute to enhancing mobility and independence for users navigating unfamiliar environments.
{
    \small
    \bibliographystyle{unsrt}
    \bibliography{main}
}


\end{document}